\documentclass[letterpaper, 10 pt, conference]{ieeeconf}
\IEEEoverridecommandlockouts
\usepackage{cite}
\usepackage{amsmath,amssymb,amsfonts}
\usepackage{graphicx}
\usepackage{textcomp}
\usepackage{xcolor}
\usepackage{hyperref}
\usepackage{multirow}
\usepackage{stfloats}
\usepackage{array}
\usepackage{booktabs}
\usepackage{adjustbox}
\usepackage{multirow}
\usepackage{placeins}
\usepackage{float}
\DeclareMathOperator*{\softmax}{softmax}

\usepackage{microtype} \emergencystretch=3em

\title{\LARGE \bf
MSCloudCAM: Multi-Scale Context Adaptation with Convolutional Cross-Attention for Multispectral Cloud Segmentation
}

\author{\authorblockN{Md Abdullah Al Mazid, Liangdong Deng, Naphtali Rishe}
\authorblockA{\textit{School of Computing and Information Sciences} \\
\textit{Florida International University}}
}

\begin{document}

\maketitle
\begin{abstract}
Clouds remain a major obstacle in optical satellite imaging, limiting accurate environmental and climate analysis. To address the strong spectral variability and the large scale differences among cloud types, we propose MSCloudCAM, a novel multi-scale context adapter network with convolution based cross-attention  tailored for multispectral and multi-sensor cloud segmentation. A key contribution of MSCloudCAM is the explicit modeling of multiple complementary multi-scale context extractors. Our formulation uses one extractor’s fine-resolution features and the other extractor’s global contextual representations enabling dynamic, scale-aware feature selection. Building on this idea, we design a new convolution-based cross attention adaptation module (CAM) that effectively fuses localized, detailed information with broader multi-scale context rather than simply stacking or concatenating context extractors' outputs. Integrated with a hierarchical vision backbone and refined through channel and spatial attention mechanisms, MSCloudCAM achieves strong spectral–spatial discrimination. Experiments on multisensor datasets e.g. CloudSEN12 (Sentinel-2) and L8Biome (Landsat-8), demonstrate that MSCloudCAM achieves superior overall segmentation performance and competitive class-wise accuracy compared to recent state-of-the-art models, while maintaining efficient model complexity, highlighting the effectiveness of the proposed design for large-scale Earth observation.

\end{abstract}

\section{Introduction}
Clouds significantly obscure the Earth’s surface in optical satellite imagery, creating persistent challenges for remote sensing applications. Accurate detection and classification of cloud types are essential for atmospheric correction, land surface monitoring, and environmental modeling.  

Recent deep learning–based cloud masking and segmentation methods have shown strong performance by leveraging hierarchical feature representations. However, most existing approaches struggle to generalize across heterogeneous sensors and spectral configurations. This limitation becomes especially critical in multi-class cloud segmentation requires modeling subtle spectral differences and large variations in spatial scale.

\begin{figure}[t]
    \centering
    \includegraphics[width=0.9\linewidth]{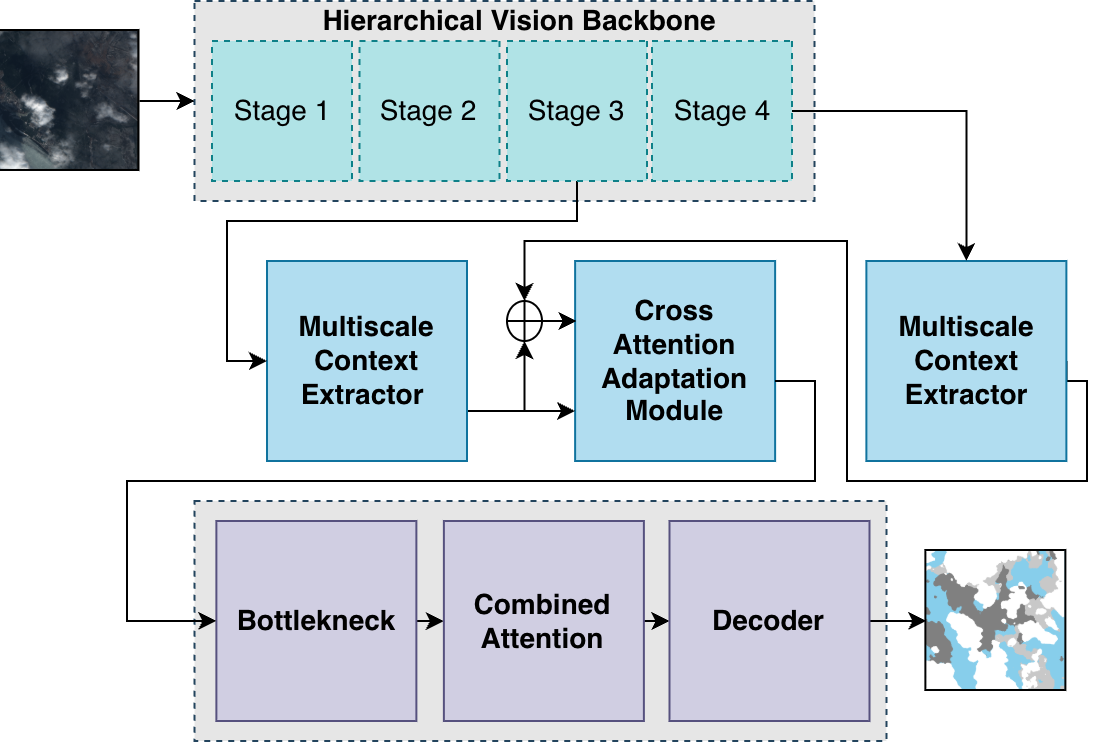}
    \caption{High-level architecture of MSCloudCAM. The model integrates a hierarchical vision backbone with dual multi-scale context extractors and introduces a novel convolutional cross-attention adaptation module (CAM).}
    \label{fig:mscloudcam-arch}
\end{figure}  

To address these challenges, we propose MSCloudCAM, a cross-attention adapter with multi-scale context network designed for robust cloud segmentation in multispectral, multi-sensor imagery. Our key contributions are as follows:

\begin{itemize}
    \item We introduce a \textbf{hybrid multi-scale context framework that leverages two complementary multi-scale context extractors}. One emphasizes on fine-grained spatial details and the other on capturing broad global semantics. Thus it enables us to achieve comprehensive multi-scale feature utilization capability from a hierarchical vision backbone.
    
    \item We design a \textbf{novel convolution-based cross-attention adaptation module, i.e. CAM} that leverages these complementary feature hierarchies through an adaptive formulation. This mechanism enables dynamic, scale-aware feature fusion, offering a more principled alternative to conventional stacking or concatenation.
    
    \item We demonstrate that MSCloudCAM achieves superior performance on both Sentinel-2 (CloudSEN12) \cite{aybar2022cloudsen12} and Landsat-8 (L8Biome) \cite{foga2017cloud}, consistently outperforming state-of-the-art models while maintaining competitive computational efficiency.
\end{itemize}

\section{Related Work}

\subsection{Cloud Segmentation in Remote Sensing}
Cloud segmentation is essential for atmospheric correction, land cover mapping, and environmental monitoring. Early rule-based methods such as Fmask \cite{zhu2014automated} relied on band thresholds and decision trees but frequently misclassified thin clouds, mixed pixels, and bright surfaces. Spectral index–based approaches (e.g., NDSI, NDVI, cloud probability) \cite{hollstein2016ready} provided improved robustness yet still struggled under strong spectral variability or semi-transparent clouds.  
Traditional machine learning classifiers (SVM, Random Forests, boosting) \cite{singh2022cloud} incorporated spectral–textural cues but remained constrained by handcrafted features and limited context modeling, motivating the shift toward deep learning.

\subsection{Deep Learning Approaches}
CNN-based deep learning architectures such as UNet \cite{ronneberger2015u} and {UNetMobv2} \cite{sandler2018mobilenetv2} capture fine localization through skip connections.DeepLabV3+ \cite{chen2018encoder} enhances multi-scale context via dilated convolutions and ASPP. Numerous CNN variants target cloud detection specifically, including CDNetv1/v2 \cite{yang2019cdnet, guo2020cdnetv2}, KappaMask \cite{domnich2021kappamask}, DBNet \cite{lu2022dual}, SCNN \cite{chai2024remote}, MCDNet \cite{dong2024mcdnet}, and HRCloudNet \cite{li2024high}, each improving spectral–spatial fusion or multi-scale representation.

Transformer-based methods leverage self-attention for long-range dependency modeling, crucial for large or spatially diffuse clouds. SegFormer \cite{xie2021segformer} uses a lightweight MiT backbone for global context, while Mask2Former \cite{cheng2022masked} delivers strong generalization through masked attention. UPerNet-InternImage \cite{wang2023internimage} combines hierarchical transformers with pyramid pooling to achieve state-of-the-art results in remote sensing segmentation.

Hybrid models merge CNNs for low-level edge and texture extraction with transformer blocks for global reasoning, including HRNet–Transformer variants and cross-attention designs \cite{liu2021swin}.  

Foundation-model–based segmentation adapts pretrained vision transformers such as DINOv2 \cite{oquab2024dinov} to remote sensing; however, effective use of multispectral data still requires spectral alignment and domain-specific fine-tuning.

\subsection{Gaps in Current Methods}
Despite significant progress, most approaches are designed for single-sensor datasets, limiting cross-sensor generalization. Moreover, few models explicitly integrate complementary multi-scale spectral–spatial cues using directional cross-attention adaptation. Our MSCloudCAM addresses both challenges.

\section{Methodology}

\subsection{Overview}
\begin{figure*}[t]
    \centering
    \includegraphics[width=\linewidth]{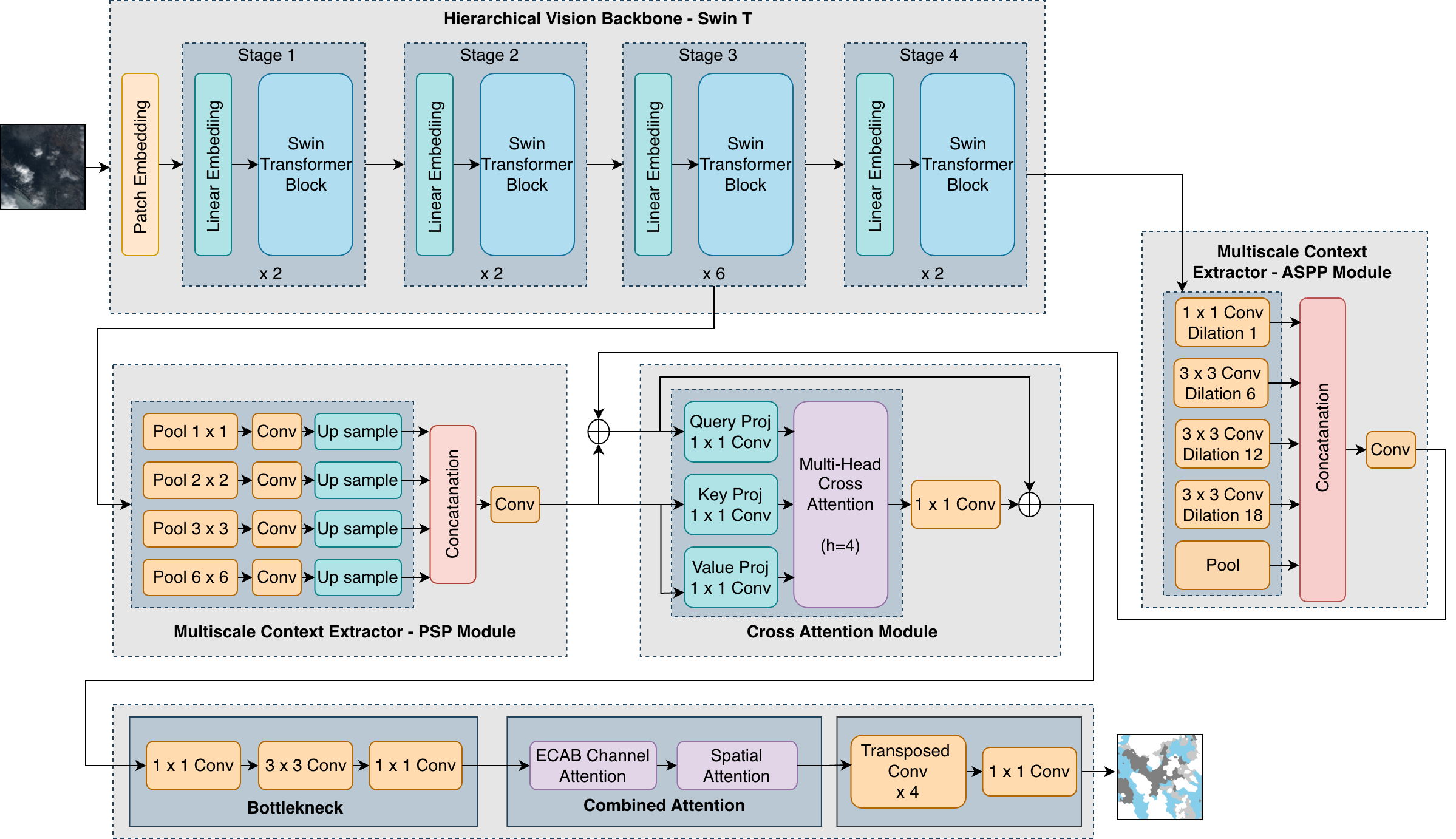}
    \caption{Architecture of MSCloudCAM. A multispectral Swin Transformer encoder generates hierarchical features. ASPP and PSP modules enrich deep and intermediate representations. These are fused by convolutional cross-attention adapter, refined via channel–spatial attention, and decoded by a multi-stage upsampling path with auxiliary supervision.}
    \label{fig:architecture}
\end{figure*}

MSCloudCAM is a multispectral semantic segmentation network that integrates a hierarchical vision backbone e.g. Swin Transformer \cite{liu2021swin} encoder with dual context extractor modules, a cross-attention adaptation mechanism, and a lightweight (only adds 1.7M trainable parameters) combined attention refinement block. The model is designed to capture long-range spectral–spatial dependencies while retaining fine-grained local structure. A multi-stage decoder with auxiliary supervision produces the final dense predictions. The overall pipeline is illustrated in Fig.~\ref{fig:architecture}.

\subsection{Backbone Feature Extraction}
Let $\mathbf{X}\in\mathbb{R}^{B\times C_{\text{in}}\times H_0\times W_0}$ denote the input multispectral image with $C_{\text{in}}$ spectral channels (13 for Sentinel-2, 11 for Landsat-8), where $B$ is the batch size and $(H_0, W_0)$ denote the input spatial resolution. 
The encoder $\mathcal{E}(\cdot)$ produces a hierarchy of features:
\begin{equation}
    \{f_1,f_2,f_3,f_4\} = \mathcal{E}(\mathbf{X}), \quad f_i \in \mathbb{R}^{B\times C_i \times H_i \times W_i},
\end{equation}
where $(H_i, W_i)$ denote the spatial resolutions at stage $i$, progressively reduced from $(H_0, W_0)$ through hierarchical downsampling, and $C_i \in \{96, 192, 384, 768\}$. 
Shifted-window self-attention within Swin ensures that $f_4$ captures global dependencies, while $f_1$–$f_3$ preserve finer spatial details.

\subsection{Multi-Scale Context Modules}
To enhance representational richness, we employ two complementary multi context encoders that operate at different feature hierarchies.

As part of the dual multiscale context extractor, ASPP \cite{chen2018encoder} is applied to the deepest feature map $f_4$. 
Atrous Spatial Pyramid Pooling expands the effective receptive field through parallel dilated convolutions with multiple rates and an additional global context branch:
\begin{align}
    x_{\text{ASPP}} &= \phi_{\text{ASPP}}(f_4) \nonumber \\
    &= \bigoplus_{r\in\{1,6,12,18\}} 
       \sigma\!\big(\text{Conv}_{3\times3}^{(r)}(f_4)\big) 
       \;\oplus\; \sigma\!\big(\text{GAP}(f_4)\big),
\end{align}
where $\oplus$ denotes channel-wise concatenation, $r$ specifies the dilation rate, $\text{GAP}$ is global average pooling, and $\sigma$ represents the ReLU nonlinearity. 
This operation aggregates contextual information across multiple receptive fields.

For the intermediate feature map $f_3$, Pyramid Scene Parsing (PSP) \cite{zhao2017pyramid} aggregates contextual cues by adaptive pooling at several grid scales:
\begin{equation}
    x_{\text{PSP}} = \phi_{\text{PSP}}(f_3) = 
    \bigoplus_{s\in\{1,2,3,6\}} 
    \sigma\!\left(\text{Up}\big(\text{Pool}_s(f_3)\big)\right),
\end{equation}
where where $s\in\{1,2,3,6\}$ denotes the pooling grid size and $\text{Pool}_s$ maps $f_3$ to an $s\times s$ grid via average pooling, followed by projection and bilinear upsampling. This operation encodes multi-scale scene layout information.

This asymmetric placement aligns contextual operators with the feature hierarchy. 
PSP is applied to $f_3$ to exploit its higher spatial resolution for preserving structural details and accurate boundary localization, whereas ASPP is applied to the deeper $f_4$ where stronger semantic abstractions and a coarser resolution benefit more from large-dilation convolutions and global context aggregation.

\subsection{Cross-Attention Adaptation}
The outputs from the two multi-scale context extractors are concatenated to form a joint feature representation:
\begin{equation}
    x_{\text{cat}} = [\,x_{\text{MC1}} \,\|\, x_{\text{MC2}}\,] 
    \in \mathbb{R}^{B \times (C_a+C_p)\times H \times W},
\end{equation}
where $x_{\text{MC1}}$ emphasizes fine spatial detail from ASPP module and $x_{\text{MC2}}$ from PSP module encodes broader contextual semantics.  
To effectively combine these complementary cues, we employ a convolutional cross-attention adaptation mechanism.  
Here, $x_{\text{cat}}$ is used to generate the \emph{queries}, as the concatenated representation benefits from selectively incorporating global context to refine local structure.  
Meanwhile, $x_{\text{MC2}}$ provides the \emph{keys} and \emph{values}, since its pooled multi-scale features offer stable, region-level guidance suitable for modulating finer details.  
This directional interaction allows each spatial location in $x_{\text{MC1}}$ to attend to the most relevant semantic context encoded by $x_{\text{MC2}}$.

Formally, the attention is computed as follows: the query $Q = W_Q * x_{\text{cat}}$, key $K = W_K * x_{\text{MC2}}$, and value $V = W_V * x_{\text{MC2}}$ are first obtained.
\begin{align}
    \text{Attn}(Q,K,V) 
      &= \softmax\!\Big(\tfrac{QK^\top}{\sqrt{d_k}}\Big)V, \\
    \tilde{x} &= W_O * \text{Attn}(Q,K,V) + x_{\text{cat}},
\end{align}
where $*$ denotes convolution and residual addition preserves the original spatial representation. This formulation enables fine-grained features to be adaptively enhanced by global contextual cues, producing a richer and more discriminative multi-scale representation.

\subsection{Bottleneck Projection}
The fused representation $\tilde{x}$ is compressed and regularized through a three-layer bottleneck:
\begin{align}
z &= \phi_{\text{bottleneck}}(\tilde{x}) \nonumber \\
  &= \sigma\!\Big(\text{Conv}_{1\times1}\!\big(
        \sigma(\text{Conv}_{3\times3}\!(
        \sigma(\text{Conv}_{1\times1}(\tilde{x}))))\big)\Big),
\end{align}
which first reduces dimensionality, then captures local spatial dependencies, and finally projects features into a compact 512-channel embedding. This step reduces redundancy and prepares the representation for attention-based refinement.

\subsection{Combined Attention Refinement}
The bottleneck output is recalibrated by a combined attention operator:
\begin{equation}
    z' = \phi_{\text{CA}}(z) = \phi_{\text{ECA}}(z) \odot \phi_{\text{SA}}(z) + z,
\end{equation}
where $\phi_{\text{ECA}}$ denotes Efficient Channel Attention \cite{wang2020eca} and $\phi_{\text{SA}}$ denotes spatial attention \cite{woo2018cbam, liu2022spatial}.  
This joint formulation first reweights channels based on their global importance and then emphasizes spatially salient regions, allowing the model to focus on both discriminative spectral responses and structurally meaningful areas.

\subsection{Decoder with Deep Supervision}
The refined representation $z'$ is progressively upsampled through transposed convolutions:
\begin{equation}
    y_{i} = \sigma(\text{Deconv}_i(y_{i-1})), \quad i=1,\dots,4,
\end{equation}
with $y_0=z'$. Two intermediate outputs, $\hat{Y}_1$ and $\hat{Y}_2$, are generated from $y_1$ and $y_2$, respectively, and resized to the input resolution $(H_0,W_0)$. The final prediction $\hat{Y}$ is obtained from $y_4$. The training objective jointly supervises all three outputs:
\begin{equation}
    \mathcal{L} = \lambda_{\text{final}}\mathcal{L}(\hat{Y},Y) + \lambda_{1}\mathcal{L}(\hat{Y}_1,Y) + \lambda_{2}\mathcal{L}(\hat{Y}_2,Y),
\end{equation}
where $Y$ is the ground truth mask, and $(\lambda_{\text{final}},\lambda_1,\lambda_2)$ are supervision weights.

\begin{table*}[!ht]
\centering
\vspace{2.5mm}
\large{
\caption{Performance comparison on CloudSEN12 L1C (TOA).}
\label{tab:cloudsen12_l1c_results}
\resizebox{\textwidth}{!}{
\begin{tabular}{lccccc|ccccc|ccccc|c}
\hline
\multirow{2}{*}{Model} & \multicolumn{5}{c|}{IoU (\%)} & \multicolumn{5}{c|}{F1 (\%)} & \multicolumn{5}{c|}{Accuracy (\%)} & aAcc \\
 & Clear & Thick & Thin & Shadow & mIoU & Clear & Thick & Thin & Shadow & mF1 & Clear & Thick & Thin & Shadow & mAcc & (\%) \\
\hline
DBNet \cite{lu2022dual} & 85.20 & 43.02 & 81.40 & 52.44 & 65.52 & 92.01 & 60.16 & 89.75 & 68.80 & 77.68 & \textbf{95.56} & 53.15 & 89.40 & 62.47 & 75.15 & 86.83 \\

CDNetv2 \cite{guo2020cdnetv2} & 84.98 & 43.37 & 80.67 & 53.40 & 65.60 & 91.88 & 60.50 & 89.30 & 69.62 & 77.83 & 94.80 & 53.13 & 89.32 & 65.34 & 75.65 & 86.68 \\
HRCloudNet \cite{li2024high} & 86.01 & 45.88 & \textbf{83.54} & 57.61 & 68.26 & 92.48 & 62.90 & \textbf{91.03} & 73.10 & 79.88 & 94.49 & 59.47 & 91.39 & 67.19 & 78.13 & 87.86 \\
UNet-MobV2 & 90.82 & 85.98 & 50.54 & 67.21 & 73.64 & 95.19 & 92.46 & 67.14 & 80.39 & 83.80 & 94.42 & 96.11 & 95.05 & 96.62 & 95.55 & 91.10 \\
DeepLabV3+-MobV2 & 90.13 & 85.75 & 48.97 & 66.29 & 72.79 & 94.81 & 92.33 & 65.75 & 79.73 & 83.15 & 94.02 & 96.04 & 94.53 & 96.51 & 95.28 & 90.55 \\
UPerNet-InternImage & 90.85 & 85.28 & 36.70 & 68.87 & 70.42 & 95.09 & 91.82 & 47.94 & 80.98 & 78.96 & 94.50 & {96.43} & 95.24 & 96.85 & 95.75 & 91.51 \\
Mask2Former-DINOv2 & 88.50 & 82.10 & 43.61 & 63.62 & 69.46 & 93.90 & 90.17 & 60.73 & 77.76 & 80.64 & 92.94 & 94.84 & 94.41 & 96.11 & 94.57 & 89.15 \\
Mask2Former-Swin-T & 90.75 & 86.70 & 49.90 & 69.53 & 74.22 & 95.15 & 92.88 & 66.58 & 82.02 & 84.16 & 94.46 & 96.32 & 94.34 & {96.93} & 95.51 & 91.03 \\
SegFormer-MiT-B5 & 91.47 & 86.70 & 49.77 & {70.24} & 74.54 & {95.55} & 92.88 & 66.46 & {82.52} & 84.35 & 94.84 & 96.27 & 95.25 & 96.92 & 95.82 & 91.64 \\
MSCloudCAM & \textbf{91.64} & \textbf{87.24} & {52.58} & \textbf{70.63} & \textbf{75.52} & \textbf{95.64} & \textbf{93.19} & {68.92} & \textbf{82.78} & \textbf{85.13} & {94.93} & \textbf{96.42} & \textbf{95.58} & \textbf{97.04} & \textbf{95.99} & \textbf{91.99} \\
\hline
\end{tabular}
}
}
\end{table*}

\begin{table*}[!ht]
\centering
\vspace{2.5mm}
\large{
\caption{Performance comparison on {CloudSEN12 L2A (BOA)}.}
\label{tab:cloudsen12_l2a_results}
\resizebox{\textwidth}{!}{
\begin{tabular}{lccccc|ccccc|ccccc|c}
\hline
\multirow{2}{*}{Model} & \multicolumn{5}{c|}{IoU (\%)} & \multicolumn{5}{c|}{F1 (\%)} & \multicolumn{5}{c|}{Accuracy (\%)} & aAcc \\
 & Clear & Thick & Thin & Shadow & mIoU & Clear & Thick & Thin & Shadow & mF1 & Clear & Thick & Thin & Shadow & mAcc & (\%) \\
\hline

DBNet \cite{lu2022dual} & 85.42 & 42.76 & 80.80 & 53.62 & 65.65 & 92.14 & 59.90 & 89.38 & 69.81 & 77.81 & 94.83 & 53.53 & 90.27 & 63.45 & 75.52 & 86.82 \\
CDNetv2 \cite{guo2020cdnetv2} & 85.35 & 44.01 & 80.85 & 54.00 & 66.05 & 92.09 & 61.12 & 89.41 & 70.13 & 78.19 & 95.65 & 56.37 & 88.06 & 63.56 & 75.91  & 86.88 \\
HRCloudNet \cite{li2024high} & 87.24 & 44.18 & \textbf{82.78} & 59.20 & 68.35 & 93.19 & 61.28 & \textbf{90.58} & 74.37 & 79.85 & 95.41 & 51.00 & 93.17 & 69.83 & 77.35 & 88.35 \\
UNet-MobV2 & 88.74 & 83.87 & 54.13 & 60.15 & 71.72 & 94.03 & 91.23 & 70.24 & 75.12 & 82.65 & 93.45 & 95.55 & 93.71 & 95.43 & 94.53 & 89.07 \\
DeepLabV3+-MobV2 & 88.01 & 83.70 & 50.36 & 61.78 & 70.96 & 93.62 & 91.12 & 66.98 & 76.37 & 82.03 & 92.93 & 95.46 & 93.83 & 95.45 & 94.42 & 88.84 \\
UPerNet-InternImage & 87.79 & 82.22 & 40.31 & 63.96 & 68.57 & 93.30 & 89.78 & 51.63 & 77.19 & 77.98 & 93.08 & 95.55 & 93.48 & 95.50 & 94.40 & 88.81 \\
Mask2Former-DINOv2 & 83.53 & 79.32 & 45.51 & 56.50 & 66.21 & 91.03 & 88.47 & 62.55 & 72.21 & 78.56 & 90.30 & 94.23 & 91.17 & 94.93 & 92.66 & 85.31 \\
Mask2Former-Swin-T & 88.91 & \textbf{84.35} & 50.72 & 63.25 & 71.81 & 94.13 & \textbf{91.51} & 67.30 & 77.49 & 82.61 & 93.61 & \textbf{95.77} & 93.33 & 95.30 & 94.50 & 89.01 \\
SegFormer-MiT-B5 & \textbf{89.28} & 83.50 & 51.87 & \textbf{64.19} & 72.21 & \textbf{94.34} & 91.01 & 68.31 & \textbf{78.19} & 82.96 & \textbf{93.76} & 95.36 & 93.86 & \textbf{95.90} & 94.72 & 89.44 \\
MSCloudCAM & \textbf{89.28} & {84.30} & 55.53 & {63.38} & \textbf{73.12} & {94.33} & {91.48} & 71.40 & {77.58} & \textbf{83.70} & \textbf{93.76} & {95.69} & \textbf{94.20} & {95.74} & \textbf{94.85} & \textbf{89.69} \\
\hline
\end{tabular}
}
}
\end{table*}
\subsection{Datasets}

We evaluate MSCloud on two publicly available multispectral cloud segmentation datasets.

CloudSEN12 \cite{aybar2022cloudsen12} is a global Sentinel-2 cloud segmentation dataset 
with 13 spectral bands (10–60\,m) and pixel-level labels for \emph{clear} (0), 
\emph{thick cloud} (1), \emph{thin cloud} (2), and \emph{cloud shadow} (3). 
We use the \emph{high} subset, selecting 10k samples each from L1C (Top-of-Atmosphere - TOA) and L2A ( Bottom-of-Atmosphere - BOA). All bands are normalized by dividing top-of-atmosphere reflectance by $3000.0$.

L8Biome \cite{foga2017cloud} is derived from Landsat-8 Cloud Cover Assessment, containing 11 bands (30\,m) with manual masks. The original five classes—\emph{fill} (0), \emph{shadow} (64), \emph{clear} (128), 
\emph{thin cloud} (192), and \emph{cloud} (255)—are remapped to CloudSEN12’s taxonomy: clear $\rightarrow 0$, thick cloud $\rightarrow 1$, thin cloud $\rightarrow 2$, shadow $\rightarrow 3$. The \emph{fill} class is ignored. Bands are normalized with the same $3000.0$ divisor for consistency.

\begin{table}[h]
\centering
\caption{Dataset split statistics for CloudSEN12 and L8Biome.}
\label{tab:dataset_split}
\begin{tabular}{lccc}
\hline
Dataset & Train & Validation & Test \\
\hline
CloudSEN12 L1C & 8,500 & 500 & 1,000 \\
CloudSEN12 L2A & 8,500 & 500 & 1,000 \\
L8Biome & 13,378 & 4,459 & 4,459 \\
\hline
\end{tabular}
\end{table}

\subsection{Loss Function}
The training objective is a weighted sum of the losses from the final prediction 
and the two auxiliary predictions:
\begin{equation}
\mathcal{L} = \lambda_{\text{final}}\mathcal{L}_{\text{final}} +
              \lambda_{1}\mathcal{L}_{1} +
              \lambda_{2}\mathcal{L}_{2}.
\end{equation}
We set $\lambda_{\text{final}}=1.0$ and $\lambda_1=\lambda_2=0.4$, 
ensuring the final prediction dominates while auxiliary supervision 
stabilizes early-stage training.

\subsection{Training Protocol}

Our model is trained using the Adam optimizer with an initial learning rate of $1\times10^{-4}$. The batch size is set to 8, and we train for 100 epochs.

\subsection{Evaluation Metrics}

\begin{align}
\mathrm{IoU}_i &= \frac{\mathrm{TP}_i}{\mathrm{TP}_i+\mathrm{FP}_i+\mathrm{FN}_i}, \\[6pt]
\mathrm{F1}_i  &= \frac{2\,\mathrm{TP}_i}{2\,\mathrm{TP}_i+\mathrm{FP}_i+\mathrm{FN}_i}, \\[6pt]
\mathrm{Acc}_i &= \frac{\mathrm{TP}_i+\mathrm{TN}_i}{\mathrm{TP}_i+\mathrm{FP}_i+\mathrm{FN}_i+\mathrm{TN}_i}.
\end{align}

\noindent
Here, $\mathrm{TP}_i$, $\mathrm{FP}_i$, $\mathrm{FN}_i$, and $\mathrm{TN}_i$ denote the numbers of true positives, false positives, false negatives, and true negatives for class $i$, respectively. Class-wise metrics ($\mathrm{IoU}_i$, $\mathrm{F1}_i$, $\mathrm{Acc}_i$) evaluate individual categories, while their means (mIoU, mF1, mAcc) are obtained by averaging across $K$ classes. In contrast, overall accuracy (aAcc) is computed over all pixels.

\section{Results}
\subsection{Quantitative Performance}
We first evaluate the models quantitatively in terms of IoU, F1, and Accuracy on three benchmark datasets (CloudSEN12 L1C, CloudSEN12 L2A, and L8Biome). 
Tables~\ref{tab:cloudsen12_l1c_results}, \ref{tab:cloudsen12_l2a_results}, \ref{tab:l8biome_results} present the detailed results.
\begin{table*}[!ht]
\centering
\vspace{2.5mm}
\large{
\caption{Performance comparison on L8Biome.}
\label{tab:l8biome_results}
\resizebox{\textwidth}{!}{
\begin{tabular}{lccccc|ccccc|ccccc|c}
\hline
\multirow{2}{*}{Model} & \multicolumn{5}{c|}{IoU (\%)} & \multicolumn{5}{c|}{F1 (\%)} & \multicolumn{5}{c|}{Accuracy (\%)} & aAcc \\
 & Clear & Thick & Thin & Shadow & mIoU & Clear & Thick & Thin & Shadow & mF1 & Clear & Thick & Thin & Shadow & mAcc & (\%) \\
\hline
DBNet \cite{lu2022dual} & 80.19 & 38.18 & \textbf{82.37} & 04.90 & 51.41 & 89.00 & 55.25 & \textbf{90.34} & 09.34 & 60.99 & \textbf{96.51} & 49.34 & 87.89 & 05.03 & 59.70 & 83.62 \\

CDNetv2 \cite{guo2020cdnetv2} & 73.81 & 24.85 & 75.85 & 00.00 & 43.63 & 84.93 & 39.81 & 86.27 & 00.00 & 52.75 & 90.18 & 31.95 & \textbf{89.81} & 00.00 & 52.98 & 78.56 \\

HRCloudNet \cite{li2024high} & 72.41 & 30.89 & 70.76 & 00.00 & 43.51 & 83.99 & 47.20 & 82.88 & 00.00 & 53.52 & 85.50 & 43.70 & 85.88 & 00.00 & 53.77 & 77.04 \\

UNet-MobV2 & \textbf{88.98} & 78.59 & {39.97} & 6.37 & 53.48 & {94.17} & 88.01 & {57.11} & 11.98 & 62.82 & {94.26} & 91.50 & 87.55 & 98.55 & 92.97 & 85.93 \\

DeepLabV3+-MobV2 & 87.25 & 79.21 & 33.37 & 2.54 & 50.59 & 93.19 & 88.40 & 50.05 & 4.96 & 59.15 & 93.32 & 91.25 & 87.50 & 98.54 & 92.65 & 85.31 \\

UPerNet-InternImage & 54.68 & 46.97 & 18.65 & 1.56 & 30.46 & 60.19 & 53.40 & 26.72 & 2.55 & 35.72 & 91.00 & 90.58 & 83.73 & 98.35 & 90.92 & 81.83 \\
Mask2Former-DINOv2 & 69.67 & 70.75 & 34.67 & 13.13 & 47.06 & 82.12 & 82.87 & 51.49 & {23.22} & 59.93 & 84.03 & 86.86 & 84.39 & {98.65} & 88.48 & 76.96 \\
Mask2Former-Swin-T & 84.36 & 78.64 & 37.50 & 13.54 & 53.51 & 91.52 & 88.04 & 54.55 & 23.85 & 64.49 & 91.74 & 91.62 & 86.01 & 98.62 & 92.00 & 84.00 \\
SegFormer-MiT-B5 & {88.62} & 74.38 & 39.52 & 15.27 & {54.45} & \textbf{93.97} & 85.31 & 56.65 & {26.50} & {65.61} & 94.04 & 90.12 & 86.05 & \textbf{98.70} & 92.23 & 84.45 \\
MSCloudCAM & {88.27} & \textbf{82.23} & {40.19} & \textbf{21.48} & \textbf{58.04} & {93.77} & \textbf{90.25} & {57.34} & \textbf{35.36} & \textbf{69.18} & {93.84} & \textbf{92.91} & {88.74} & \textbf{98.70} & \textbf{93.55} & \textbf{87.09} \\
\hline
\end{tabular}
}
}
\end{table*}

\subsection{Qualitative Results}
\begin{figure*}[t]
  \centering
  \includegraphics[width=\textwidth]{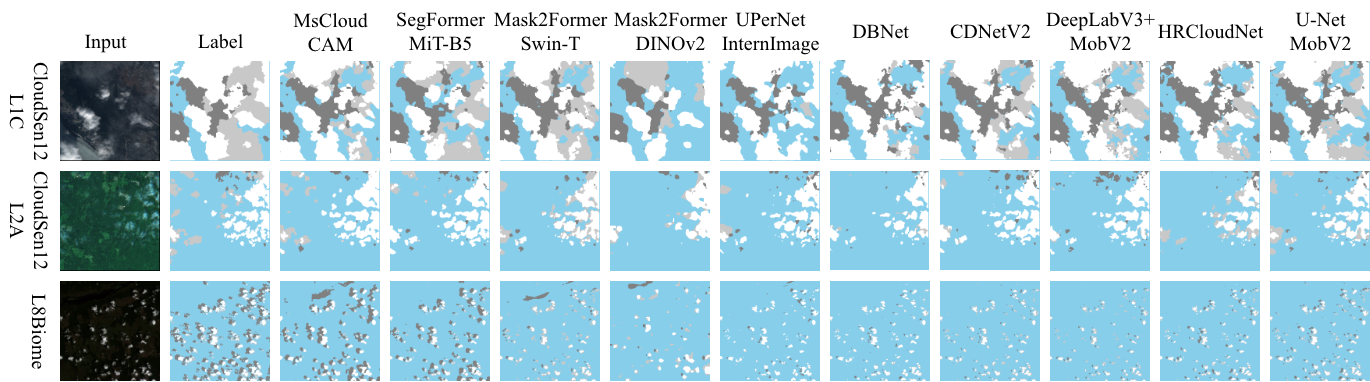}
  \caption{Qualitative cloud segmentation results across datasets. Each row represents a dataset.The proposed method shows sharper delineation of thin clouds and shadows than other approaches. Class–color mapping: Class 1 (Clear sky – light blue), Class 2 (Thick cloud – white), Class 3 (Thin cloud – light gray), Class 4 (Cloud shadow – dark gray)}
  \label{fig:qual_results}
\end{figure*}

To further demonstrate the effectiveness of our method, Fig.~\ref{fig:qual_results} presents visual comparisons of cloud segmentation results.

\subsection{Ablation Study}
To evaluate the role of each multicontext component, we conduct ablation experiments by selectively replacing or removing modules. Replacing ASPP with PSP enhances global context but loses fine detail, while using ASPP instead of PSP improves multi-scale local features but sacrifices global scene understanding. 

Using PSP and ASPP together provides complementary global–local context and outperforms either module alone. Removing the cross-attention module (CAM) and using simple concatenation weakens the model’s ability to emphasize informative regions. The full MSCloudCAM system with PSP, ASPP, and CAM delivers the highest accuracy. Overall, these findings confirm that each component contributes uniquely to the system.

\begin{table}[!ht]
    \centering
    \caption{Ablation study evaluating dual multi context extractor and cross-attention module (CAM) on CloudSEN12 L1C, CloudSEN L2A, L8Biome datasets.}
    \label{tab:ablation_dual_extractor_context_adapter}
    \begin{tabular}{l | c c | c c c}
        \hline
        \textbf{Dataset} & \textbf{Extractor} & \textbf{Fusion} &
        \textbf{mIoU} & \textbf{mF1} & \textbf{mAcc}\\
        \hline 
        CloudSEN12 & PSP +\ PSP     & CAM    & 72.31 & 83.56 & 94.11\\  
        L1C & ASPP +\ ASPP   & CAM    & 71.98 & 83.37 & 93.89\\ 
         & PSP +\ ASPP    & Concat  & 73.43 & 84.29 & 94.75\\ 
         & PSP + \ ASPP    & CAM    & \textbf{75.52} & \textbf{85.13} & \textbf{95.99}\\ 
        \hline
        CloudSEN12 & PSP +\ PSP     & CAM    & 70.26 & 81.61 & 92.05\\  
        L2A & ASPP +\ ASPP   & CAM    & 70.09 & 81.25 & 91.92\\ 
         & PSP +\ ASPP    & Concat  & 71.48 & 82.44 & 92.63\\ 
         & PSP + \ ASPP    & CAM    & \textbf{73.12} & \textbf{83.70} & \textbf{94.85}\\
        \hline
        L8Biome & PSP +\ PSP     & CAM    & 54.95 & 67.63 & 91.87\\  
         & ASPP +\ ASPP   & CAM    & 55.08 & 67.31 & 91.46\\ 
         & PSP +\ ASPP    & Concat  & 56.41 & 68.22 & 92.02\\ 
         & PSP + \ ASPP    & CAM    & \textbf{58.04} & \textbf{69.18} & \textbf{93.55}\\ 

        \hline
    \end{tabular}
\end{table}

\subsection{Model Complexity Analysis}
Table~\ref{tab:model_complexity} summarizes the computational complexity of evaluated models in terms of FLOPs, parameter counts, and inference efficiency. In addition to latency (ms/image), we report throughput (TP, samples/s) to quantify batch inference performance. All experiments were conducted on a single NVIDIA A100 40GB GPU using PyTorch 2.3 and CUDA 12.1, with a batch size of 8 and an input resolution of 512×512. All models were evaluated in FP32 precision. Our proposed MSCloudCAM achieves a noteworthy balance between accuracy and computational efficiency and CAM only adds 1.7M trainable parameters to the model.

\begin{table}[!ht]
\centering
\vspace{2.5mm}
\caption{Model complexity comparison in terms of FLOPs (G), parameters (M), inference time (ms/image), and TP as throughput (FPS).}
\label{tab:model_complexity}
\begin{tabular}{lcccc}
\hline
\textbf{Model} & \textbf{Flops$\downarrow$} & \textbf{Params$\downarrow$} & \textbf{Latency$\downarrow$} & \textbf{TP$\uparrow$} \\
\hline
U-Net-MobileNetV2 & 13.6 & 6.6 & 11.5 & 87.0 \\
DeepLabV3+-MobV2 & 6.1 & 4.3 & 14.8 & 67.5 \\
DeepLabV3+-ResNet101 & 56.1 & 45.6 & 38.5 & 26.0 \\
UPerNet-InternImage & 193 & 110 & 52.0 & 19.2 \\
Mask2Former-Swin-T & 73.09 & 47.4 & 33.6 & 29.7 \\
Mask2Former-DINOv2 & 85.6 & 90.6 & 44.0 & 22.7 \\
SegFormer-MiT-B5 & 55.4 & 45.6 & 28.7 & 34.8 \\
\textbf{MSCloudCAM} & \textbf{38.98} & \textbf{47.44} & \textbf{15.9} & \textbf{62.7} \\
\hline
\end{tabular}
\end{table}

\section{Code and Model Availability}
The full implementation of our proposed MSCloudCAM framework is publicly available. The source code, training scripts, and experimental configurations are hosted on GitHub: \url{https://github.com/mazid-rafee/ms-cloudcam}. Additionally, pretrained model checkpoints are released on Hugging Face: \url{https://huggingface.co/mazid-rafee/MS-CloudCAM/tree/main}.

\section{Conclusion}
We presented MSCloudCAM which integrates hierarchical vision backbone with dual multi-scale context extractors and employs our novel convolution based cross attention adaptation module. Extensive experiments on the CloudSEN12 and L8Biome datasets demonstrate that MSCloudCAM consistently surpasses state-of-the-art methods across multiple segmentation metrics. These highlight the effectiveness and generalizability of our design for multi-sensor and multi-spectral cloud understanding. While our experiments focus on transformer-based backbones, evaluating the proposed context adapter with CNN-based encoders remains an interesting direction for future work.

\section*{Acknowledgment}
This material is based in part upon work supported by the National Science Foundation under Grants CNS-2018611 and CNS-1920182.

\bibliographystyle{IEEEtran}
\bibliography{references}

\end{document}